\documentclass[runningheads]{llncs}
\usepackage[T1]{fontenc}
\usepackage{graphicx}
\usepackage{latexsym}
\usepackage{amssymb}
\usepackage{amsmath}
\usepackage{booktabs}
\usepackage{enumitem}
\usepackage{graphicx}
\usepackage{color}
\usepackage{array}
\usepackage{multirow}
\usepackage[symbol]{footmisc}

\usepackage{pgfplots}
\pgfplotsset{compat=1.12}
\usepgfplotslibrary{statistics}
\usepgfplotslibrary{colormaps}
\usepackage{tikz}
\usetikzlibrary{arrows, matrix, positioning, math, external, patterns}

\usepackage[binary-units=true, detect-all]{siunitx}[=v2]
\DeclareSIUnit{\noop}{\relax}
\pgfplotsset{yticklabel = {\SI[round-mode=figures]{\tick}{\noop}}}

\begin{document}
\title{Taking a SEAT: Predicting Value Interpretations from Sentiment, Emotion, Argument, and Topic Annotations}
\titlerunning{Predicting Value Interpretations from SEAT Annotations}
%
\author{Adina Nicola Dobrinoiu\inst{*}
\and Ana Cristiana Marcu\inst{*}
\and Amir Homayounirad
\and Luciano Cavalcante Siebert
\and Enrico Liscio
}
\authorrunning{A.N. Dobrinoiu, A.C. Marcu et al.}
%
\institute{Delft University of Technology, the Netherlands\\
\email{\{A.N.Dobrinoiu, A.C.Marcu-1\}@student.tudelft.nl}\\
\email{\{A.Homayounirad, L.CavalcanteSiebert, E.Liscio\}@tudelft.nl}
}
\maketitle              
\begin{abstract}
Our interpretation of value concepts is shaped by our sociocultural background and lived experiences, and is thus subjective. Recognizing individual value interpretations is important for developing AI systems that can align with diverse human perspectives and avoid bias toward majority viewpoints. To this end, we investigate whether a language model can predict individual value interpretations by leveraging multi-dimensional subjective annotations as a proxy for their interpretive lens. That is, we evaluate whether providing examples of how an individual annotates Sentiment, Emotion, Argument, and Topics (SEAT dimensions) helps a language model in predicting their value interpretations. Our experiment across different zero- and few-shot settings demonstrates that providing all SEAT dimensions simultaneously yields superior performance compared to individual dimensions and a baseline where no information about the individual is provided. Furthermore, individual variations across annotators highlight the importance of accounting for the incorporation of individual subjective annotators. To the best of our knowledge, this controlled setting, although small in size, is the first attempt to go beyond demographics and investigate the impact of annotation behavior on value prediction, providing a solid foundation for future large-scale validation.

\end{abstract}
\section{Introduction}

\footnotetext[1]{Equal contribution.}
\renewcommand*{\thefootnote}{\arabic{footnote}}

Human values are deeply intertwined with argumentation, influencing how people form opinions and justify their stances \cite{Bench-Capon2003,kobbe2020exploring}. Much of real-world argumentation implicitly appeals to values as commonly accepted reasons why a position is desirable or ``right'' \cite{kiesel2022identifying,rokeach1973nature}. Being able to recognize these underlying values is crucial for NLP applications such as argument mining, persuasive dialogue systems, and aligning AI responses with human morality \cite{liscio2023value}. 

Predicting which values a given text or argument expresses is an inherently subjective endeavor, due to the subjective nature of valuing \cite{mackie1988subjectivity}. Different individuals can interpret the same text differently, possibly reflecting their own values and perspectives. This phenomenon is a specific instance of perspectivism, a broader challenge in the AI community: many language understanding tasks lack a single ``ground truth'' label because meaning is partly in the eye of the beholder. Recent perspectivist approaches argue that we should embrace, rather than erase, such differences \cite{creanga2024designing}. Instead of leaning into majority aggregation or discarding disagreements (which obscures the rich diversity of valid interpretations), perspectivism advocates that AI models should be capable of recognizing and modeling each individual’s viewpoint \cite{cabitza2023toward}. 

In this work, we investigate whether language models can predict and represent the diversity of interpretation of the values behind a piece of text. In other words, can a language model detect that two individuals might read the same text differently? 
Addressing this question is important for developing AI that is sensitive to whom it is interacting with, with implications for personalized AI assistants and fairness (ensuring minority viewpoints are not ignored by models).

A common approach in prior research has been to use demographic attributes as proxies for human differences. For instance, gender, age, or culture are often used to tailor models or prompts, on the assumption that these attributes correlate with the person’s judgments \cite{Fleisig2023WhenTM,goyal2022toxicity,Wan2023}. However, recent findings show that even explicitly adding annotators’ demographic information into a model did not significantly improve its ability to predict those annotators’ labels \cite{orlikowski2023ecological}. 
Similarly, attempts to prompt language models with demographic personas (e.g., asking the model to respond ``as a 30-year-old woman'') have yielded limited accuracy in simulating that group’s perspective \cite{orlikowski2025beyond}. This suggests that individual annotation behavior depends on more than broad sociodemographic categories. 

We explore a new approach to approximate an individual’s value interpretation: leveraging the person’s prior textual annotations across related dimensions. Several textual dimensions are frequently studied in NLP with values \cite{liu2012survey,mohammad2013crowdsourcing}. In particular, we use a person’s annotations for Sentiment, Emotion, Argument, and Topic, which we term the \textit{SEAT dimensions}, as a proxy for their interpretive lens. Intuitively, how someone recognizes sentiment and emotion in text, what arguments they find expressed, and which topics they focus on can reveal their underlying value interpretations. 
This approach moves beyond static demographics to behavioral cues embedded in language. Recent work lends support to this idea: \cite{jiang2024re} found that an annotator’s task-specific attitudes and preferences were more predictive of their labeling behavior than their demographic traits. In other words, knowing how a person tends to label content is a better indicator of their perspective than knowing their age or gender. 

Specifically, we investigate whether providing past SEAT annotations can guide a language model in predicting a person’s value interpretations. We frame the task as follows: given a language model and a particular individual, we provide the model with a text sample and ask it to predict how the individual would interpret the human values expressed in the text sample. We then evaluate how the prediction changes when providing different amounts of that individual’s past annotations on the SEAT dimensions for the sample text and/or other similar texts. 
To test this approach, we ask five individuals to annotate a survey dataset \cite{Itten2022} with the values and the SEAT dimensions expressed in every data point. We then employ a language model to predict the values behind the texts that compose the dataset, by providing different levels of information on each annotator's SEAT annotations. Our results show that providing a few examples of the full set of SEAT annotations improves the prediction of individual value interpretations. Instead, providing examples with the individual SEAT dimensions does not improve over the off-the-shelf baseline.

This work provides early insights into language models’ sensitivity to individual differences in value interpretation based on non-demographic textual cues.
To our knowledge, this is the first attempt to personalize value prediction by conditioning a language model on a user’s multi-dimensional annotation history, paving the way for more refined behavioral cues that can be used in support of predicting individual value interpretations, such as multi-modal and non-textual cues.
Such a nuanced approach could contribute to improved deliberation support by providing participants with counterfactual scenarios of value interpretation based on the SEAT dimensions.

\section{Related Works and Background}

We start by reviewing works on the theoretical background behind human values and their connection to the SEAT dimensions. Next, we review literature on the prediction of values in text and LLM-based approaches to it.

\subsection{Human Values and SEAT Dimensions}
\label{sec:related-works:values}

Human values are guiding principles individuals use to shape their choices and actions \cite{schwartz2012overview}. The Schwartz theory of fundamental human values categorizes values into universal dimensions such as benevolence, achievement, tradition, and autonomy, which consistently emerge across cultures as central determinants of human judgment and behavior \cite{schwartz2012overview}. These values influence how people form opinions, argue, and deliberate, shaping individual and collective decision-making processes \cite{kiesel2022identifying}. 
\cite{rokeach1973nature} further clarifies values as beliefs relating to desirable end-states or modes of conduct and describes value systems as hierarchies of values formed by cultural, social, and personal factors, emphasizing that values belong inherently to personas rather than objects, enabling systematic analysis of individual, essential for understanding human disagreement and decision making.

Several textual dimensions frequently studied in NLP have been associated with values, particularly within deliberative contexts \cite{liu2012survey,mohammad2013crowdsourcing}. \textbf{S}entiment analysis, focusing on identifying positive or negative attitudes, can serve as an indicator of implicit value judgments underlying these attitudes \cite{zhangsentiment}. \textbf{E}motion recognition categorizes explicit affective states such as joy, anger, or sadness, and it has been employed to improve the identification of the underlying values expressed in discourse \cite{jafari2024unveiling}. \textbf{A}rgument mining captures structural elements that describe why particular positions are considered desirable or persuasive, and because of their motivational nature, these elements have often been associated with values \cite{kiesel2022identifying,kiesel-etal-2023-semeval}. \textbf{T}opic modeling identifies the thematic domains emphasized in discourse, 
with the identification of topics shown to be influenced by speakers' cultural backgrounds and personal value systems
\cite{rezapour2019enhancing}. 
We collectively refer to Sentiment, Emotion, Argument, and Topic as the four \textit{SEAT dimensions}. These SEAT dimensions provide comprehensive textual proxies for exploring how values are expressed in text, which might make them suitable as auxiliary inputs to a language model when predicting individual interpretations of values.

\subsection{Value Prediction through NLP Approaches}

A traditional method for classifying values in text involves using value dictionaries---lists of words associated with specific values---by analyzing the relative frequency of these words \cite{Pennebaker2001}, such as in the Moral Foundation Dictionary \cite{Graham2013}. These dictionaries have been extended using semi-automated approaches \cite{Araque2020,Hopp2020,Wilson2018} and NLP methods \cite{Araque2021,Ponizovskiy2020}, while the limitations of word count methods have been addressed with word embedding models \cite{Bahngat2020,Garten2018,Pavan2020MoralityText}.

Recent approaches leverage supervised machine learning \cite{Alshomary2022,kiesel2022identifying,Liscio2025COLM,Liscio2023ACL,van-der-meer-etal-2023-differences}, where language models are trained on datasets with value annotations, such as ValueNet \cite{Qiu2022ValueNet:System} or Value Kaleidoscope \cite{sorensen2024value}.
The introduction of the ValueEval '23 \cite{kiesel-etal-2023-semeval} and '24 \cite{kiesel:2024d} shared task has cemented value classification into a recognized challenge, with most participants employing the supervised learning paradigm.

However, supervised learning requires the availability of human annotations, which may be unfeasible in domain-specific applications such as debating energy transition policies in a Dutch municipality (i.e., the dataset we use in our experiments, as detailed in Section~\ref{sec:dataset}). The transfer learning paradigm has been evaluated but proven to be of limited efficacy for the value prediction task \cite{Huang2022,Liscio2022a}. To this end, recent approaches have shown promising value prediction results with the zero-shot prompting paradigm with LLMs \cite{mishra2024eric,senthilkumar2024leveraging}, which we employ in this work to predict values in text and extend by employing the SEAT dimensions as auxiliary information to recognize individuals' value interpretations.

\subsection{Prompting Large Language Models}

State-of-the-art language models have demonstrated generalization capability through prompting techniques such as zero-shot and few-shot learning \cite{wei2022emergent}. In a zero-shot setting, models receive only instructions and generalize to new tasks without task-specific fine-tuning, whereas few-shot prompts guide models by providing a handful of illustrative examples in-context \cite{brown2020language}. Building on this, recent methods have attempted to steer model outputs toward individual styles or viewpoints by prompting the model to adopt a specific persona---for example, instructing it to respond as ``a friendly and outgoing person'' or using descriptions of a user’s characteristics \cite{zhang2024personalization}. However, simplistic persona-based prompts, often reliant on demographic attributes, have shown limited success in capturing individuals' nuanced and diverse preferences \cite{dong2024can,orlikowski2023ecological}. Personalization based on behavioral rather than demographic signals may align better with actual user preferences and lead to improved alignment between model predictions and individual annotations \cite{jiang2024re}. Our study builds on this hypothesis by using SEAT annotations as auxiliary input when prompting a language model to predict the values expressed in a piece of text, testing the model's capacity to reflect individual users' value judgments based on their annotation behaviors.

\section{Methodology and Experiments}

We frame value prediction as a multi-label text classification task: given an input text, the goal is to identify which human values (from a predefined set) are expressed or implied (based on an annotator's interpretation of the text). We hypothesize that incorporating context from related subjective NLP tasks can improve value identification. In particular, we leverage annotations from \textbf{S}entiment analysis, \textbf{E}motion recognition, \textbf{A}rgument detection, and \textbf{T}opic classification as additional inputs. By providing these auxiliary signals about the text’s sentiment, emotional tone, argumentative content, or topic, we expect the language model to better infer the expressed values. The key question is whether an LLM can use annotations from these related tasks to more accurately predict human values (e.g., does knowing the text’s emotional tone or topic help in determining if it reflects values like honesty or success?). We evaluate this hypothesis by comparing the model’s value predictions with and without such auxiliary annotations. Data and code are publicly available\footnote{https://github.com/adina-dobrinoiu/SEAT}.

\subsection{Data and Annotations}
\label{sec:dataset}

We experiment with the Energy Transition PVE dataset \cite{Itten2022}, a survey administered to citizens of the Súdwest-Fryslân municipality in the Netherlands to support in co-creating an energy transition policy. Within the survey, citizens could distribute their votes among several options and write textual \textit{justifications} for their choices. These justifications provide insight into the values that motivate them, which can be identified through NLP methods as shown by \cite{liscio2025value}.

We select a subset of 50 justifications and ask five annotators to annotate each of them with the following five dimensions: (1) \textit{Sentiment}---the overall sentiment polarity of the justification on a five-point scale from Very Negative to Very Positive, (2) \textit{Emotion}---the prominent emotions expressed (one or multiple from a popular emotion taxonomy \cite{demszky-etal-2020-goemotions}, e.g. curiosity, confusion), (3) \textit{Argument}---a summary or key snippet of the argumentative content in the justification (if any), (4) \textit{Topic}---a brief descriptor of the topic or domain of the justification (e.g. energy independence, social engagement), and, finally, (5) \textit{Values}---the human values described in the Schwartz value theory, adapted by \cite{kiesel2022identifying}. The value annotation was performed with the original 54 labels, which we grouped into the 20 parent labels to make the experiments and analysis manageable, in line with previous work \cite{kiesel-etal-2023-semeval,kiesel:2024d}. Appendix~\ref{appendix:annotation} reports the demographics of the annotators, the exact instructions, and the task labels.
%


Table~\ref{table:annotator_agreement} shows the inter-annotator agreement for the five annotation dimensions. Fleiss' kappa is used for sentiment, emotion, topic, and values because they involve selecting predefined categorical labels. Argument annotation requires identifying text spans that may vary in length and position, which makes categorical agreement measures like Fleiss’ kappa unsuitable. Instead, agreement is measured using the pairwise $F_1$-score, which treats one annotator’s spans as ground truth and another’s as predictions, averaged over all annotator pairs. We observe that agreement is high for topic but relatively low for the other dimensions, underscoring the need for a personalized approach.

\begin{table}[h]
\centering
\caption{Inter-annotator agreement for the annotation dimensions, measured through Fleiss' kappa except for argument, which is measured through pairwise $F_1$-score.}
\label{table:annotator_agreement}
\begin{tabular}{
  @{}
  >{\centering}p{1.8cm}
  >{\centering}p{1.8cm}
  >{\centering}p{1.8cm}
  >{\centering}p{1.8cm}
  >{\centering\arraybackslash}p{1.8cm}
  @{}
}
\toprule
\textbf{Sentiment} & \textbf{Emotion} & \textbf{Argument} & \textbf{Topic} & \textbf{Values} \\
\midrule
0.17 & 0.00365 & 0.2447 & 0.514 & 0.0144\\
\bottomrule
\end{tabular}
\end{table}

\subsection{Model and Prompting Strategy}
\label{sec:prompts}

We use a pretrained 8-billion-parameter language model to perform value prediction via prompted generation (meta-llama/Meta-Llama-3.1-8B-Instruct). We use an in-context learning paradigm---the model is provided with a prompt that includes task instructions, optional examples, and one textual justification (which we refer to as the \textit{reference justification}), and asked to identify the value label(s) expressed in this reference justification. Appendix~\ref{Prompt Design} describes the exact prompts. Precisely, we start by testing the following baseline:

\begin{description}[leftmargin=0em,listparindent=1em,itemsep=.5em]
\item[\textbf{Zero-shot baseline (ZS)}] 
The language model is provided with the reference justification and a list of values to choose from, and asked to identify the values expressed in the justification. We employ this as our baseline, where the model is provided no information about the individual.
\end{description}

Next, we evaluate whether providing additional information on an individual's annotation behavior leads to a more accurate prediction of the value that the individual considers to be expressed in the reference justification. To test this, for each annotator and each reference justification, we provide the model with information on how the annotator annotated the reference justification and other justifications with the SEAT dimensions.

\begin{description}[leftmargin=0em,listparindent=1em,itemsep=.5em]
\item[\textbf{One-shot prompt (OS)}] 
We extend the ZS baseline by providing the language model with the information on how the individual annotated the reference justification with one (or more) SEAT dimension.

\item[\textbf{Few-shot prompt (FS)}] 
We extend the OS prompt by providing additional examples of how the annotator annotated other justifications with one (or more) SEAT dimensions. To ensure that the few-shot examples in the prompt are relevant to the reference justification’s content, we compute sentence embeddings using a pretrained SentenceTransformer model (all-MiniLM-L6-v2) and perform a K-nearest neighbors search in the embedding space to find the $K$ most semantically similar justifications to the reference justifications in the dataset. We experiment with different values of $K$, as detailed in the next subsection.

\end{description}

\subsection{Experimental Setup}
\label{sec:exp-setup}


We evaluate the approaches described in Section~\ref{sec:prompts} in the following settings:

\begin{description}[leftmargin=0em,listparindent=1em,itemsep=.5em]
\item[\textbf{Dimension-specific}] 
The model is given only one SEAT dimension annotation at a time in the prompt. We run separate experiments for each SEAT dimension to analyze which auxiliary signal is most informative. For example, in the \textit{emotion} setting for the OS prompt, for each annotator and each justification, the prompt provides as auxiliary information the emotion that the annotator identified in the reference justification (and likewise, the few-shot prompts include the emotion annotations for each additionally provided example).

\item[\textbf{All dimensions}] 
The model is given all four SEAT dimensions simultaneously to evaluate whether a combined set of contextual information is more effective than the single ones.
For example, in the \textit{all} setting for the OS prompt, for each annotator and each justification, the prompt provides as auxiliary information all the SEAT dimensions that the annotator identified in the reference justification (and likewise, the few-shot prompts include all these annotations for each additionally provided example). 
\end{description}

We test the OS and FS prompts with each of the SEAT dimensions and with a combination of the dimensions. In the FS prompts, we experiment with $K=4$, $K=9$, and $K=14$, thus resulting in a total of 5, 10, and 15 provided examples (including the reference justification under evaluation). Table~\ref{table:settings} provides an overview of the resulting 20 settings we test in addition to the ZS baseline.

\begin{table}[h]
\centering
\caption{Overview of the settings tested for each annotator.}
\label{table:settings}
\begin{tabular}{{@{}>{}p{2.2cm}@{}>{\centering}p{1.8cm}@{}>{\centering}p{1.8cm}@{}>{\centering}p{1.8cm}@{}>{\centering}p{1.8cm}@{}>{\centering\arraybackslash}p{1.8cm}@{}}}
\toprule
\textbf{Method} & \textbf{Sentiment} & \textbf{Emotion} & \textbf{Argument} & \textbf{Topic} & \textbf{All} \\
\midrule
One-shot & OS-S & OS-E & OS-A & OS-T & OS-all \\
Few-shot (5) & FS-5-S & FS-5-E & FS-5-A & FS-5-T & FS-5-all \\
Few-shot (10) & FS-10-S & FS-10-E & FS-10-A & FS-10-T & FS-10-all\\
Few-shot (15) & FS-15-S & FS-15-E & FS-15-A & FS-15-T & FS-15-all\\
\bottomrule
\end{tabular}
\end{table}


For each of the 21 settings and each annotator, we perform value prediction for all 50 justifications in the dataset. We repeat this for the five annotators, resulting in 21x5=105 experiments (with each experiment consisting of performing value prediction on the full dataset). We repeat each experiment five times with five different seeds to account for variability, and consider a value as predicted when it appears in at least three of the five repetitions. We report the average micro $F_1$-score as the metric for model performance.


\section{Results}

Table~\ref{table:all-results} shows the experimental results across the different methods, settings, and annotators. 

\begin{table}[h!]
\centering
\caption{Complete overview of the results ($F_1$-score) for the five annotators (represented by their IDs). In bold, the best results per annotator; underlined, the results that are not significantly different from the best result.}
\label{table:all-results}
\begin{tabular}{@{}>{\centering}p{0.8cm}@{}>{}p{2.2cm}@{}>{\centering}p{1.8cm}@{}>{\centering}p{1.8cm}@{}>{\centering}p{1.8cm}@{}>{\centering}p{1.8cm}@{}>{\centering\arraybackslash}p{1.8cm}@{}}
\toprule
\textbf{ID} & \textbf{Method} & \textbf{Sentiment} & \textbf{Emotion} & \textbf{Argument} & \textbf{Topic} & \textbf{All} \\
\midrule
\multirow{5}{*}{1} & One-shot & 0.180 & 0.155 & 0.137 & \underline{0.229} & \underline{0.219} \\
& Few-shot (5) & 0.160 & 0.189 & 0.174 & 0.138 & \textbf{0.302} \\
& Few-shot (10) & 0.118 & 0.173 & 0.169 & 0.153 & \underline{0.291} \\
& Few-shot (15) & 0.166 & 0.160 & 0.145 & 0.146 & \underline{0.275} \\
\cmidrule{2-7}
& Baseline & \multicolumn{5}{c}{0.164} \\
\midrule
\midrule
\multirow{5}{*}{2} & One-shot & 0.210 & 0.282 & 0.185 & 0.284 & 0.246 \\
& Few-shot (5) & 0.263 & 0.244 & 0.245 & 0.217 & 0.314 \\
& Few-shot (10) & 0.249 & 0.234 & 0.264 & 0.207 & \textbf{0.382} \\
& Few-shot (15) & 0.254 & 0.218 & 0.252 & 0.192 & \underline{0.380} \\
\cmidrule{2-7}
& Baseline & \multicolumn{5}{c}{0.244} \\
\midrule
\midrule
\multirow{5}{*}{3} & One-shot & 0.182 & 0.184 & 0.181 & 0.207 & 0.190 \\
& Few-shot (5) & 0.200 & \underline{0.229} & \underline{0.222} & 0.199 & \underline{0.244} \\
& Few-shot (10) & 0.204 & \underline{0.234} & 0.207 & 0.200 & \textbf{0.298} \\
& Few-shot (15) & \underline{0.210} & \underline{0.223} & 0.194 & 0.184 & \underline{0.241} \\
\cmidrule{2-7}
& Baseline & \multicolumn{5}{c}{0.230} \\
\midrule
\midrule
\multirow{5}{*}{4} & One-shot & 0.261 & 0.246 & 0.164 & \underline{0.301} & 0.252 \\
& Few-shot (5) & 0.226 & \underline{0.266} & 0.240 & 0.199 & \underline{0.335} \\
& Few-shot (10) & 0.234 & 0.263 & 0.258 & 0.221 & \textbf{0.339} \\
& Few-shot (15) & 0.234 & 0.260 & 0.229 & 0.214 & \underline{0.309} \\
\cmidrule{2-7}
& Baseline & \multicolumn{5}{c}{0.221} \\
\midrule
\midrule
\multirow{5}{*}{5} & One-shot & 0.257 & 0.228 & 0.215 & 0.287 & 0.271 \\
& Few-shot (5) & 0.306 & 0.261 & 0.189 & 0.293 & \underline{0.406} \\
& Few-shot (10) & 0.343 & 0.336 & 0.301 & 0.312 & \textbf{0.444} \\
& Few-shot (15) & 0.312 & 0.329 & 0.343 & 0.300 & \underline{0.428} \\
\cmidrule{2-7}
& Baseline & \multicolumn{5}{c}{0.228} \\
\bottomrule
\end{tabular}
\end{table}

The best results are consistently achieved by providing all SEAT dimensions through the few-shot paradigm. To facilitate further interpretation of the results, in the next subsections, we break down specific aspects of these results to highlight the main findings.

\subsection{Providing All SEAT Annotations Helps}
\label{sec:providing-all-SEAT}

\begin{figure}[h]
\centering
\begin{tikzpicture}
\begin{axis}[
width=\columnwidth,
height=5cm,
ylabel=$F_1$-score,
xlabel=Auxiliary information,
ybar,
xtick={0,1,2,3,4},
xticklabels={Sentiment,Emotion,Argument,Topic,All},
ymin=0,
ymax=0.45,
bar width=6pt,
enlarge x limits=0.1,
legend pos=north west,
legend cell align={left},
legend columns=5,
legend style={/tikz/every even column/.append style={column sep=0.3cm}},
ymajorgrids=true,
grid style=dashed,
]

\addplot[blue,dashed,sharp plot,update limits=false, line width=1.5] coordinates { (-1,0.217) (6,0.217) };
\addplot [pink,fill=pink!30!white,postaction={pattern=horizontal lines}] table[x={x}, y={1}] {tikz/seat_k.dat};
\addplot [teal,fill=teal!30!white,postaction={pattern=vertical lines}] table[x={x}, y={5}] {tikz/seat_k.dat};
\addplot [violet,fill=violet!30!white,postaction={pattern=dots}] table[x={x}, y={10}] {tikz/seat_k.dat};
\addplot [orange,fill=orange!30!white,postaction={pattern=north east lines}] table[x={x}, y={15}] {tikz/seat_k.dat};

\legend{ZS,OS,FS-5,FS-10,FS-15}
\end{axis}
\end{tikzpicture}
\caption{$F_1$-score resulting from including different auxiliary information in the prompt, averaged over the five annotators. The horizontal dashed line displays the ZS baseline ($F_1=0.217$) averaged over the five annotators.}
\label{fig:auxiliary-info}
\end{figure}
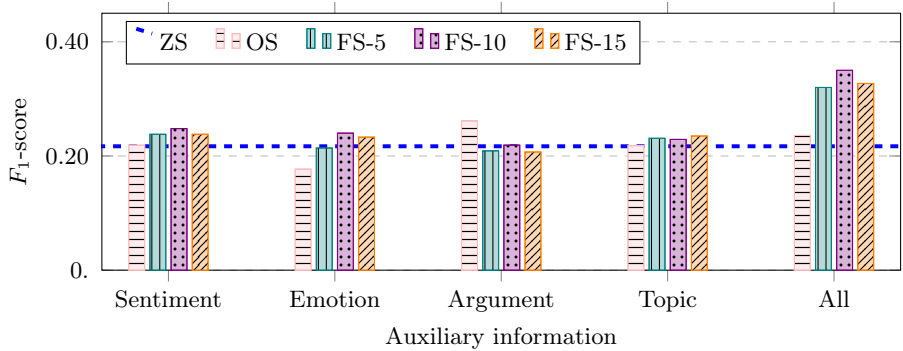

We begin by investigating how providing different types of auxiliary information impacts the prediction of individual value interpretations. Fig.~\ref{fig:auxiliary-info} shows the results grouped by the auxiliary information that is provided in the prompt (averaged over the five annotators). 

We observe that, in line with Table~\ref{table:all-results}, providing a few examples with all SEAT annotations improves the prediction of individual value interpretations. Instead, providing examples of the individual SEAT annotations does not improve results over the ZS baseline.
This suggests that value interpretation is a complex cognitive process that benefits from multiple contextual cues. Rather than relying on a single dimension, the model appears to integrate information across the four dimensions when inferring underlying values. This finding aligns with psychological theories of value-based decision-making, which emphasize the multi-faceted nature of moral and value judgment \cite{ajzen1991theory,haidt2001emotional,rokeach1973nature}.

\subsection{Similarities and Differences across Individuals}

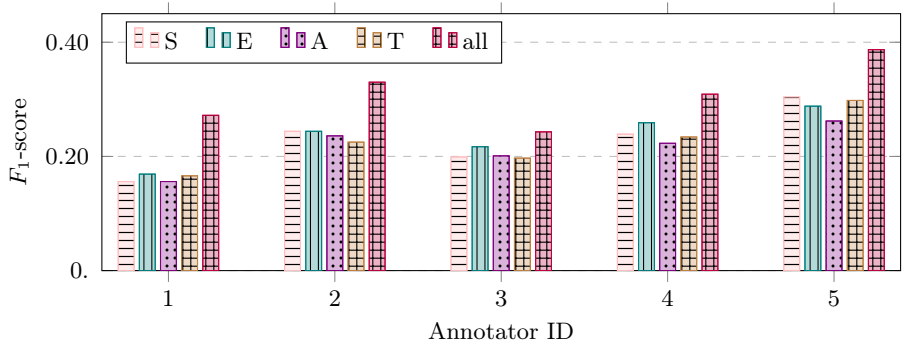
\begin{figure}[h]
\centering
\begin{tikzpicture}
\begin{axis}[
width=\columnwidth,
height=5cm,
ylabel=$F_1$-score,
xlabel=Annotator ID,
ybar,
xtick={0,1,2,3,4},
xticklabels={1,2,3,4,5},
ymin=0,
ymax=0.45,
bar width=6pt,
enlarge x limits=0.1,
legend pos=north west,
legend cell align={left},
legend columns=5,
legend style={/tikz/every even column/.append style={column sep=0.3cm}},
ymajorgrids=true,
grid style=dashed,
]
\addplot [pink,fill=pink!30!white,postaction={pattern=horizontal lines}] table[x={x}, y={S}] {tikz/annotators_seat.dat};
\addplot [teal,fill=teal!30!white,postaction={pattern=vertical lines}] table[x={x}, y={E}] {tikz/annotators_seat.dat};
\addplot [violet,fill=violet!30!white,postaction={pattern=dots}] table[x={x}, y={A}] {tikz/annotators_seat.dat};
\addplot [brown,fill=brown!30!white,postaction={pattern=grid}] table[x={x}, y={T}] {tikz/annotators_seat.dat};
\addplot [purple,fill=purple!30!white,postaction={pattern=grid}] table[x={x}, y={all}] {tikz/annotators_seat.dat};
\legend{S,E,A,T,all}
\end{axis}
\end{tikzpicture}
\caption{$F_1$-score by annotator, averaged over the different methods used to provide auxiliary information (i.e., the four rows in Table~\ref{table:settings}).}
\label{fig:annotators_and_seat}
\end{figure}

\begin{figure}[h]
\centering
\begin{tikzpicture}
\begin{axis}[
width=\columnwidth,
height=5cm,
ylabel=$F_1$-score,
xlabel=Annotator ID,
ybar,
xtick={0,1,2,3,4},
xticklabels={1,2,3,4,5},
ymin=0,
ymax=0.45,
bar width=6pt,
enlarge x limits=0.1,
legend pos=north west,
legend cell align={left},
legend columns=5,
legend style={/tikz/every even column/.append style={column sep=0.3cm}},
ymajorgrids=true,
grid style=dashed,
]
\addplot [brown,fill=brown!30!white,postaction={pattern=grid}] table[x={x}, y={0}] {tikz/annotators_k.dat};
\addplot [pink,fill=pink!30!white,postaction={pattern=horizontal lines}] table[x={x}, y={1}] {tikz/annotators_k.dat};
\addplot [teal,fill=teal!30!white,postaction={pattern=vertical lines}] table[x={x}, y={5}] {tikz/annotators_k.dat};
\addplot [violet,fill=violet!30!white,postaction={pattern=dots}] table[x={x}, y={10}] {tikz/annotators_k.dat};
\addplot [orange,fill=orange!30!white,postaction={pattern=north east lines}] table[x={x}, y={15}] {tikz/annotators_k.dat};
\legend{ZS,OS,FS-5,FS-10,FS-15}
\end{axis}
\end{tikzpicture}
\caption{$F_1$-score by annotator, averaged over the different provided auxiliary information (i.e., the five columns in Table~\ref{table:settings}).}
\label{fig:annotators_and_k}
\end{figure}
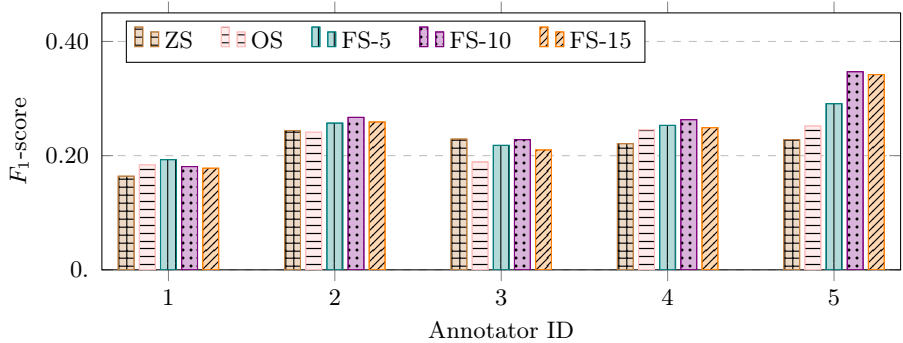

Next, we examine how the observed results differ across individuals. Fig.~\ref{fig:annotators_and_seat} and \ref{fig:annotators_and_k} show the results for each annotator, averaged over the different methods described in Section~\ref{sec:prompts} (i.e., the rows of Table~\ref{table:settings}) and over the different settings described in Section~\ref{sec:exp-setup} (i.e., the columns of Table~\ref{table:settings}), respectively.

Fig.~\ref{fig:annotators_and_seat} shows that providing all SEAT dimensions consistently improves results across all annotators. These findings are in line with the results discussed in Section~\ref{sec:providing-all-SEAT}---providing the individual SEAT annotation leads to consistently lower results, without noticeable difference among the SEAT dimensions.

Fig.~\ref{fig:annotators_and_k} shows that providing more examples generally does not improve the prediction of individual value interpretations, confirming that the type of auxiliary information has more impact than the number of examples. In fact, for all annotators, the best results are achieved with FS-5 or FS-10, rather than FS-15. We conjecture that this is due to the strategy used to select examples in the FS method---that is, the $K$ semantically most similar examples to the reference justification. A larger $K$ (e.g., in FS-15) introduces less similar examples and thus more noise in the prompt, whereas smaller values of $K$ appear to provide sufficient targeted information.
This has significant practical implications---this moderate data requirement makes the approach feasible for real-world deployment without requiring extensive historical annotation data from each annotator. This is particularly important for applications where user engagement may be limited or where privacy concerns restrict data collection. 


Finally, both figures show differences across annotators. Annotator 5 consistently achieves the highest performance across methods and settings, while Annotator 1 shows the most modest improvement.
These individual differences likely reflect varying levels of annotation consistency, familiarity with value concepts, and the degree to which personal value systems are reflected in SEAT annotations. Annotator 5's better performance may be attributed to their greater familiarity with value theory, as noted in Appendix~\ref{appendix:annotation}, leading to more systematic and predictable annotation patterns.

\subsection{Changes in Prediction}

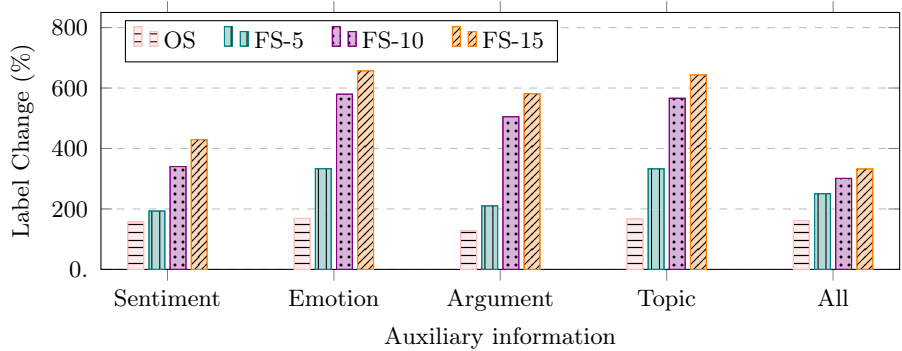
\begin{figure}[h]
\centering
\begin{tikzpicture}
\begin{axis}[
    width=\columnwidth,
    height=5cm,
    ylabel=Label Change (\%),
    xlabel=Auxiliary information,
    ybar,
    xtick={0,1,2,3,4},
    xticklabels={Sentiment,Emotion,Argument,Topic,All},
    ymin=0,
    ymax=850,  
    bar width=6pt,
    enlarge x limits=0.1,
    ymajorgrids=true,
    grid style=dashed,
    legend pos=north west,
    legend cell align={left},
    legend columns=5,
    legend style={/tikz/every even column/.append style={column sep=0.3cm}},
    yticklabel style={
        /pgf/number format/fixed,
        /pgf/number format/precision=0,
    },
]

\addplot [pink,fill=pink!30!white,postaction={pattern=horizontal lines}] table[x={x}, y={1}] {tikz/label_change_k.dat};     
\addplot [teal,fill=teal!30!white,postaction={pattern=vertical lines}] table[x={x}, y={5}] {tikz/label_change_k.dat};      
\addplot [violet,fill=violet!30!white,postaction={pattern=dots}] table[x={x}, y={10}] {tikz/label_change_k.dat};          
\addplot [orange,fill=orange!30!white,postaction={pattern=north east lines}] table[x={x}, y={15}] {tikz/label_change_k.dat}; 

\legend{OS,FS-5,FS-10,FS-15}
\end{axis}
\end{tikzpicture}
\caption{Predicted label change compared to the zero-shot ZS baseline, measured as the symmetric difference $(|A \Delta B| / |A|)$ percentage, averaged over the five annotators.}
\label{fig:label-change}
\end{figure}

The $F_1$-scores reported so far only reveal cases in which the model prediction, due to the provided auxiliary information, changes from wrong to correct (or vice versa). However, cases in which the model prediction changes without changing its correctness are also insightful, as they reveal that providing auxiliary information still prompts the model to change its prediction.
Fig.~\ref{fig:label-change} shows the changes in prediction, measured as the relative symmetric difference between baseline and alternative approach: $|A \triangle B| / |A|$, where $A$ is the set of labels predicted with the ZS baseline setting, and $B$ is the set predicted using auxiliary information. The numerator, $|A \triangle B|$, represents the number of elements that are in either $A$ or $B$, but not in both, indicating the total number of changed labels.


Surprisingly, the \textit{all} setting produces the smallest percentage of prediction change while achieving the highest $F_1$-scores. This result appears to suggest that, when the model has access to comprehensive contextual information, it makes fewer but more accurate changes to its predictions. 
This may be because individual dimensions may introduce more random variation in prediction, whereas combined dimensions provide clear guidance on when changes are warranted. In contrast, individual dimensions show higher prediction change rates but no performance improvement, suggesting that incomplete contextual information may lead to less discriminating model behavior. Nevertheless, further investigation is required to confirm these conjectures, as we elaborate next.


\section{Discussion and Limitations}

Overall, the main observed trend is that providing a more varied set of information improves individual value interpretation prediction across annotators, and is more impactful than providing just a single source of information, even if of increased quantity.
This suggests that predicting individual value interpretations requires capturing multiple aspects of an individual's interpretative style, rather than relying on a single demographic or behavioral indicator, challenging the prevailing paradigm of demographic-based personalization in NLP \cite{Fleisig2023WhenTM,goyal2022toxicity,Wan2023}. 

\paragraph{Individual variations}
The differences between annotators highlight the importance of individual-level modeling rather than assuming uniform population responses. The better results with annotator 5, who had greater familiarity with value theory, suggests that annotation expertise and conceptual clarity may be important factors in predicting personalized value interpretations. This raises questions about whether personalization works better for users with more systematic or well-articulated value systems.
Moreover, variation across individuals suggests that some users may benefit more from personalized systems than others. This could create new forms of digital inequality where users with more ``learnable'' patterns receive better AI assistance. 

\paragraph{Low performance}
The observed results are far from perfect, peaking at an $F_1$-score of 0.444. On the one hand, this speaks of the inherent difficulty in predicting subjective, individual value interpretations. On the other hand, this may suggest that the SEAT dimensions alone may be insufficient proxies for the full spectrum of human value diversity.
Additional dimensions may help capture the richness of human value prediction more effectively---among others, moral dilemmas such as trolley-type scenarios that reveal utilitarian-versus-deontological leanings \cite{marcus1980moral}; Moral Foundation scores (care, Fairness, Loyalty, etc.) obtained with the MFQ \cite{Graham2013}; Big Five personality traits, whose agreeableness and openness facets systematically predict value priorities \cite{john2010handbook}; and national cultural value profile (e.g. Hofstede's dimensions) that contextualize individual judgments within broader societal norms \cite{hofstede1980culture}. Integrating these complementary or additional variables may capture subtle facets of human value diversity and, in turn, enhance the robustness of value prediction models beyond what the SEAT dimensions alone can deliver. 

\paragraph{Generalization}
Our study used a single model architecture with specific prompting strategies. The effectiveness of our approach may vary with different model architectures, sizes, or prompting techniques. We suggest that researchers test whether larger and different models exhibit similar patterns or distinct scaling behaviors. Furthermore, the current sample size (50 data points, 5 annotators) limits generalizability; the consistency of patterns within this controlled setting provides a solid foundation for future large-scale validation.


\section{Conclusions and Future Work}

Our study demonstrates that a language model can leverage multi-dimensional subjective annotations to recognize individual value interpretations, representing a promising step toward more nuanced and individualized AI systems. The better results of combining SEAT dimensions over individual ones suggest that effective subjective value prediction requires capturing the complex interplay of multiple interpretive lenses.

This paper lays the groundwork for future research on value prediction that transcends demographic proxies to adopt more sophisticated individual modeling. While significant challenges remain, particularly in areas such as fairness and transparency, the potential benefits of more personalized and contextually aware AI systems that can predict and possibly align to individual values make this an important direction for continued investigation.


Based on our findings and limitations, we identify several directions for future research. Testing the approach with more annotators across diverse demographic and cultural backgrounds would help validate whether the synergistic effects of SEAT dimensions generalize beyond Western, educated populations and whether cultural differences in value expression affect individual value prediction. This evaluation is crucial for ensuring that personalized AI systems do not inadvertently disadvantage certain populations or create new forms of digital inequality where users with more ``learnable'' annotation patterns receive better AI assistance. 

Assessing performance across different text types and domains would provide a cross-domain evaluation system, highlighting the importance of context-specific capturing of individual value strategies. For instance, value expressions in social media posts may differ significantly from those in formal policy documents, requiring different approaches.
Multi-modal integration studies could investigate whether incorporating non-textual cues (such as voice tone, facial expressions, or physiological signals) might provide additional signals to predict value interpretations. This would be particularly relevant for applications where users interact with AI systems through multiple modalities.




\begin{credits}
\subsubsection{\ackname} Enrico Liscio's work was conducted as part of AlgoSoc, a collaborative 10-year research program on public values in the algorithmic society, and the Hybrid Intelligence Centre, both funded by the Dutch Ministry of Education, Culture and Science (OCW) under the Gravitation programme (project numbers 024.005.017 and 024.004.022). Any opinions, findings, and conclusions or recommendations expressed in this material are those of the author(s) and do not necessarily reflect the views of OCW or those of the AlgoSoc consortium as a whole.

\end{credits}
%
%
%

\clearpage

\appendix

\section{Annotation Procedure}

\label{appendix:annotation}

The annotation procedure involved five computer science students from Europe, each possessing prior knowledge in natural language processing. Among them, Annotator 5 was more familiar with the value annotation task. To ensure consistency, all annotators first reviewed the dataset and labels for each task (see Table~\ref{table:annotation-labels}). They were then instructed to examine each sentence in the dataset and assign zero or more labels from the predefined category set. Since the sentences could be interpreted in multiple ways, annotators were reminded that their judgments were inherently subjective and should reflect their individual perspectives. 

\begin{table}[h]
\centering
\caption{Overview of the annotated labels.}
\label{table:annotation-labels}
\begin{tabular}{{@{}>{}p{2cm}@{}>{}p{10cm}@{}}}
\toprule
\textbf{Category} & \textbf{Labels (comma-separated)} \\
\midrule
Sentiments & Very negative, Somewhat negative, Neutral, Somewhat positive, Very positive \\
\midrule
Emotions & admiration, amusement, anger, annoyance, approval, caring, confusion, curiosity, desire, disappointment, disapproval, disgust, embarrassment, excitement, fear, gratitude, grief, joy, love, nervousness, optimism, pride, realization, relief, remorse, sadness, surprise \\
\midrule
Arguments &  If the sentence contains an argument, the annotation consists of one or more extracted text spans corresponding to its premises (e.g., “I care about sustainability”); if no argument is present, the annotation is \textit{None}. Only premises are extracted, conclusions are not considered.\\
\midrule
Topics & Municipality and residents engagement in the energy sector, Energy storage and supplying energy in The Netherlands, Wind and solar energy, Market Determination Dynamics, Landscapes and windmills tourism, Hydrogen energy pipeline networks\\
\midrule
Values & Be creative, Be curious, Have freedom of thought, Be choosing own goals, Be independent, Have freedom of action, Have privacy, Have an exciting life, Have a varied life, Be daring, Have pleasure, Be ambitious, Have success, Be capable, Be intellectual, Be courageous, Have influence, Have the right to command, Have wealth, Have social recognition, Have a good reputation, Have a sense of belonging, Have good health, Have no debts, Be neat and tidy, Have a comfortable life, Have a safe country, Have a stable society, Be respecting traditions, Be holding religious faith, Be compliant, Be self-disciplined, Be behaving properly, Be polite, Be honoring elders, Be humble, Have life accepted as is, Be helpful, Be honest, Be forgiving, Have the own family secured, Be loving, Be responsible, Have loyalty towards friends, Have equality, Be just, Have a world at peace, Be protecting the environment, Have harmony with nature, Have a world of beauty, Be broadminded, Have the wisdom to accept others, Be logical, Have an objective view\\
\bottomrule
\end{tabular}
\end{table}

\section{Prompt Outline}
\label{Prompt Design}
Each prompt begins with a brief instruction that defines the task for the model. We prefix: “You are an expert value annotator. Your task is to extract the most relevant value labels from a given sentence.”. This primes the model to act as a value classifier.  The prompt explicitly instructs the model to output only a list of values from this predefined set and in a specific format (a Python list of strings) without any extra commentary. If auxiliary task annotations are available, the prompt includes them as additional context.
We insert a line after the sentence, saying, for instance: “Emotion Annotations for this sentence: approval, curiosity” (if using emotion context). These annotations are provided in natural language form, one per line, so the model knows the sentence’s sentiment category, emotion tags, etc. We explicitly tell the model, “You should also consider your previous annotations on the [Task] detection task,” when such context is present, to encourage it to make use of that information. Importantly, we format the prompt to resemble a few-shot learning scenario: we can include a few demonstration examples before asking the model to label the new sentence. Each example consists of a prior sentence, its annotation(s) for the auxiliary task(s), and the correct value labels for that example. For instance, in a per-task prompt, an example might look like:
Sentence: “The city reduced speed limits to improve safety.”
Sentiment Annotation: “Somewhat positive”
Values: ["Have a safe country", "Be responsible"].

We then provide the new test sentence with its annotation and ask for the Values output. In an all-tasks prompt, each example and the test input would instead list all four annotations (Argument, Emotion, Sentiment, Topic) followed by the Values for that example. By showing complete input-output examples, the model can learn the pattern that connects task annotations and sentences to the correct value labels.

\end{document}